\documentclass{article}
\usepackage[letterpaper, left=1in, right=1in, top=1in, bottom=1in]{geometry}

\usepackage{natbib}
\bibpunct{(}{)}{;}{a}{,}{,}
\usepackage[colorlinks,citecolor=blue!70!black,linkcolor=blue!70!black,
urlcolor=blue!70!black,breaklinks=true]{hyperref}

\usepackage{parskip}

\usepackage[dvipsnames]{xcolor}
\usepackage{microtype}
\usepackage{booktabs} 
\usepackage{amsmath}
\usepackage{dsfont} 
\usepackage{amssymb}
\usepackage{xspace}
\usepackage{url}            
\usepackage{algorithm}
\usepackage[noend]{algorithmic}
\usepackage{amsthm}
\usepackage{listings}
\usepackage{bm}
\usepackage{bbm}
\usepackage{enumitem}
\usepackage{nicefrac}       
\usepackage{mathrsfs} 
\usepackage{inconsolata}
\usepackage[subrefformat=parens,labelformat=parens]{subfig}
\usepackage{wrapfig}
\usepackage{graphicx}
\usepackage{bigints}
\usepackage{transparent}
\usepackage{comment}
\usepackage[nameinlink,capitalize]{cleveref}
\makeatletter
\AddToHook{cmd/appendix/before}{
\def\cref@section@alias{appendix}
\def\cref@subsection@alias{appendix}
\def\cref@subsubsection@alias{appendix}
}
\makeatother
\usepackage{thmtools}
\usepackage{thm-restate}
\usepackage{nicematrix}
\usepackage{arydshln}
\usepackage{fancyhdr}
\usepackage{mathtools}
\usepackage[scaled=.88]{helvet}
\usepackage{breakcites}
\usepackage{multirow}

\def\E{{\mathbb E}}

\DeclareMathOperator*{\argmax}{arg\,max}
\DeclareMathOperator*{\argmin}{arg\,min}

\DeclareMathOperator{\unif}{{unif}}

\renewcommand{\epsilon}{\varepsilon}

\newcommand{\algcommentlight}[1]{\textcolor{blue!70!black}{\transparent{0.5}\small{\texttt{\textbf{//\hspace{2pt}#1}}}}}

\DeclarePairedDelimiter{\brk}{[}{]}
\DeclarePairedDelimiter{\crl}{\{}{\}}
\DeclarePairedDelimiter{\prn}{(}{)}

\DeclarePairedDelimiterX{\infdiv}[2]{(}{)}{%
  #1\;\delimsize\|\;#2%
}

\newcommand{\wb}[1]{\widebar{#1}}

\def\ddefloop#1{\ifx\ddefloop#1\else\ddef{#1}\expandafter\ddefloop\fi}
\def\ddef#1{\expandafter\def\csname bb#1\endcsname{\ensuremath{\mathbb{#1}}}}
\ddefloop ABCDEFGHIJKLMNOPQRSTUVWXYZ\ddefloop
\def\ddefloop#1{\ifx\ddefloop#1\else\ddef{#1}\expandafter\ddefloop\fi}
\def\ddef#1{\expandafter\def\csname b#1\endcsname{\ensuremath{\mathbf{#1}}}}
\ddefloop ABCDEFGHIJKLMNOPQRSTUVWXYZ\ddefloop
\def\ddef#1{\expandafter\def\csname sf#1\endcsname{\ensuremath{\mathsf{#1}}}}
\ddefloop ABCDEFGHIJKLMNOPQRSTUVWXYZ\ddefloop
\def\ddef#1{\expandafter\def\csname c#1\endcsname{\ensuremath{\mathcal{#1}}}}
\ddefloop ABCDEFGHIJKLMNOPQRSTUVWXYZ\ddefloop
\def\ddef#1{\expandafter\def\csname h#1\endcsname{\ensuremath{\widehat{#1}}}}
\ddefloop ABCDEFGHIJKLMNOPQRSTUVWXYZ\ddefloop
\def\ddef#1{\expandafter\def\csname hc#1\endcsname{\ensuremath{\widehat{\mathcal{#1}}}}}
\ddefloop ABCDEFGHIJKLMNOPQRSTUVWXYZ\ddefloop
\def\ddef#1{\expandafter\def\csname t#1\endcsname{\ensuremath{\widetilde{#1}}}}
\ddefloop ABCDEFGHIJKLMNOPQRSTUVWXYZ\ddefloop
\def\ddef#1{\expandafter\def\csname tc#1\endcsname{\ensuremath{\widetilde{\mathcal{#1}}}}}
\ddefloop ABCDEFGHIJKLMNOPQRSTUVWXYZ\ddefloop
\def\ddefloop#1{\ifx\ddefloop#1\else\ddef{#1}\expandafter\ddefloop\fi}
\def\ddef#1{\expandafter\def\csname scr#1\endcsname{\ensuremath{\mathscr{#1}}}}
\ddefloop ABCDEFGHIJKLMNOPQRSTUVWXYZ\ddefloop

\let\oldparagraph\paragraph
\renewcommand{\paragraph}[1]{\oldparagraph{#1}}

\renewcommand{\epsilon}{\varepsilon}

\newcommand{\ldef}{\vcentcolon=}

\renewcommand{\bigm}[1]{%
  \ifcsname fenced@\string#1\endcsname
    \expandafter\@firstoftwo
  \else
    \expandafter\@secondoftwo
  \fi
  {\expandafter\amsmath@bigm\csname fenced@\string#1\endcsname}%
  {\amsmath@bigm#1}%
}

\newcommand{\DeclareFence}[2]{\@namedef{fenced@\string#1}{#2}}
\makeatother

\makeatletter
\let\save@mathaccent\mathaccent
\newcommand*\if@single[3]{%
  \setbox0\hbox{${\mathaccent"0362{#1}}^H$}%
  \setbox2\hbox{${\mathaccent"0362{\kern0pt#1}}^H$}%
  \ifdim\ht0=\ht2 #3\else #2\fi
  }
\newcommand*\rel@kern[1]{\kern#1\dimexpr\macc@kerna}
\newcommand*\widebar[1]{\@ifnextchar^{{\wide@bar{#1}{0}}}{\wide@bar{#1}{1}}}
\newcommand*\wide@bar[2]{\if@single{#1}{\wide@bar@{#1}{#2}{1}}{\wide@bar@{#1}{#2}{2}}}
\newcommand*\wide@bar@[3]{%
  \begingroup
  \def\mathaccent##1##2{%
    \let\mathaccent\save@mathaccent
    \if#32 \let\macc@nucleus\first@char \fi
    \setbox\z@\hbox{$\macc@style{\macc@nucleus}_{}$}%
    \setbox\tw@\hbox{$\macc@style{\macc@nucleus}{}_{}$}%
    \dimen@\wd\tw@
    \advance\dimen@-\wd\z@
    \divide\dimen@ 3
    \@tempdima\wd\tw@
    \advance\@tempdima-\scriptspace
    \divide\@tempdima 10
    \advance\dimen@-\@tempdima
    \ifdim\dimen@>\z@ \dimen@0pt\fi
    \rel@kern{0.6}\kern-\dimen@
    \if#31
      \overline{\rel@kern{-0.6}\kern\dimen@\macc@nucleus\rel@kern{0.4}\kern\dimen@}%
      \advance\dimen@0.4\dimexpr\macc@kerna
      \let\final@kern#2%
      \ifdim\dimen@<\z@ \let\final@kern1\fi
      \if\final@kern1 \kern-\dimen@\fi
    \else
      \overline{\rel@kern{-0.6}\kern\dimen@#1}%
    \fi
  }%
  \macc@depth\@ne
  \let\math@bgroup\@empty \let\math@egroup\macc@set@skewchar
  \mathsurround\z@ \frozen@everymath{\mathgroup\macc@group\relax}%
  \macc@set@skewchar\relax
  \let\mathaccentV\macc@nested@a
  \if#31
    \macc@nested@a\relax111{#1}%
  \else
    \def\gobble@till@marker##1\endmarker{}%
    \futurelet\first@char\gobble@till@marker#1\endmarker
    \ifcat\noexpand\first@char A\else
      \def\first@char{}%
    \fi
    \macc@nested@a\relax111{\first@char}%
  \fi
  \endgroup
}
\makeatother

\newcommand{\gemmamodel}{\text{Gemma 2B}\xspace}
\newcommand{\gptmodel}{\text{GPT-4o-mini}\xspace}

\newcommand{\llms}{\text{LLMs}\xspace}
\newcommand{\llm}{\text{LLM}\xspace}
\newcommand{\llmmath}{\mathsf{LLM}\xspace}
\newcommand{\alg}{\pi\xspace}
\newcommand{\cb}{\mathsf{CB}\xspace}

\newcommand{\similarity}{\texttt{Sim}\xspace}

\newcommand{\Reg}{\mathrm{\mathbf{Reg}}}

\newcommand{\corral}{\text{CORRAL}\xspace}
\newcommand{\oneshotwiki}{\texttt{OneShotWikiLinks-311}\xspace}
\newcommand{\oneshot}{\texttt{OneShotWikiLinks}\xspace}
\newcommand{\amazon}{\texttt{AmazonCat-13K}\xspace}
\newcommand{\spannerGreedy}{\textsf{SpannerGreedy}\xspace}
\newcommand{\flan}{\text{Flan-T5}\xspace}

\newcommand{\reg}{\Reg}

\title{Efficient Sequential Decision Making with Large Language Models}
\date{}

\author{
Dingyang Chen$^{\star}$\\
{\small\texttt{dingyang@email.sc.edu}}
\and
Qi Zhang$^{\star}$\\
{\small\texttt{qz5@cse.sc.edu}}
\and
Yinglun Zhu$^{\dagger}$\\
{\small\texttt{yzhu@ucr.edu}}
\and
~\\
{\normalsize $^\star$University of South Carolina \quad \quad  $^\dagger$University of California, Riverside}
}

\begin{document}
\maketitle
\begin{abstract}
This paper focuses on extending the success of large language models (LLMs) to sequential decision making. Existing efforts either (i) re-train or finetune LLMs for decision making, or (ii) design prompts for pretrained LLMs. The former approach suffers from the computational burden of gradient updates, and the latter approach does not show promising results. In this paper, we propose a new approach that leverages online model selection algorithms to efficiently incorporate LLMs agents into sequential decision making. Statistically, our approach significantly outperforms both traditional decision making algorithms and vanilla LLM agents. Computationally, our approach avoids the need for expensive gradient updates of LLMs, and throughout the decision making process, it requires only a small number of LLM calls. We conduct extensive experiments to verify the effectiveness of our proposed approach. As an example, on a large-scale Amazon dataset, our approach achieves more than a $6$x performance gain over baselines while calling LLMs in only $1.5$\% of the time steps.

\end{abstract}

\section{Introduction}
\label{sec:intro}

Sequential decision making addresses the problem of adapting an agent to an unknown environment, where the agent learns through a feedback loop by repeatedly receiving contexts, selecting actions, and observing feedback.
This approach has been widely applied in real-world scenarios, including recommendation systems \citep{li2010contextual, agarwal2016making}, healthcare \citep{tewari2017ads, svensson2023sequential}, 
and dialogue systems \citep{li2016deep}.
With the significant success of large language models (\llms) in natural language processing \citep{brown2020language, ouyang2022training, achiam2023gpt}, 
an important next step is to extend this success to sequential decision making and enhance applications therein.
\looseness=-1

\begin{figure}[t]
    \centering
    \includegraphics[width=0.5\linewidth]{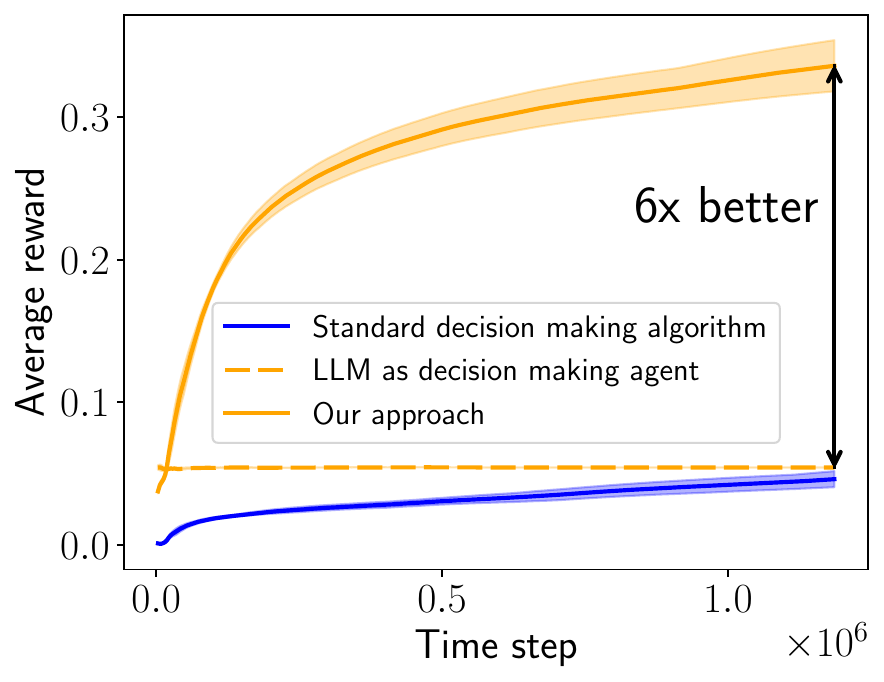}
    \caption{
    Performance comparison (higher is better) on the \amazon dataset. The decision making task is to predict item tags based on textual descriptions.
    We compare three approaches: (i) a standard decision making algorithm, (ii) a pretrained \llm as decision making agent, and (iii) our approach that balances the above two methods.
    We defer further details to \cref{sec:experiment}.}
    \label{fig:amazon}
\end{figure}

Existing efforts to leverage \llms for sequential decision making focus on two directions: 
(i) viewing decision making as sequence modeling and 
re-training or finetuning large models to adapt them to unknown environments \citep{chen2021decision, zheng2022online, reid2022can, sun2023smart, raparthy2023generalization,  lee2024supervised}, and (ii) utilizing prompt engineering and in-context learning to adapt pretrained large models to sequential decision making problems \citep{krishnamurthy2024can}.
While the first approach usually achieves promising empirical results, 
it is hindered by the substantial computational burden associated with re-training or finetuning large models, which often contain hundreds of billions of parameters.
The second approach \citep{krishnamurthy2024can}, on the other hand, 
has demonstrated that most in-context learning and prompt engineering methods fail to effectively adapt \llms to sequential decision making environments, except when employing the most advanced models (at the time), i.e., GPT-4 \citep{achiam2023gpt}, with sophisticated prompt designs.

In this paper, we propose a new approach to efficiently incorporate large pretrained models into sequential decision making environments, \emph{without the need for expensive model re-training or finetuning}.
We run experiments (see \cref{fig:amazon} and its caption for settings) on the \amazon dataset \citep{Bhatia16} and observe that:
\begin{itemize}
    \item Vanilla \llms as decision making agents exhibit strong initial performance thanks to their significant commonsense knowledge and remarkable reasoning ability. 
    However, \llm agents fail to adapt to the environment and show continuous improvements. 
    \item Standard sequential decision making algorithms, while performing poorly initially, continuously learn to adapt to the environment and improve their performance over time.
\end{itemize}

To take advantage of both methods, we adapt online model selection algorithms \citep{auer2002nonstochastic, agarwal2017corralling, pacchiano2020model} to a framework that 
can automatically balance the performance of \llm-powered policies/agents and standard decision making algorithms. Initially, the framework relies more on \llm-powered policies to achieve good initial results. 
As standard decision making algorithms begin to adapt to the environments, it gradually shifts towards these algorithms.
To our knowledge, this work presents the first result in leveraging online model selection algorithms to efficiently incorporate \llms into sequential decision making.
Our framework also offers several compelling advantages:
\begin{itemize}
    \item \textbf{Statistical efficiency.} It achieves superior performance compared to vanilla \llm-powered policies and standard sequential decision making algorithms. As shown in \cref{fig:amazon}, our approach achieves more than a $6$x performance gain ($0.336$ vs. $0.054$) compared to baselines.

    \item \textbf{Computational efficiency.}
    First, our approach does not require expensive re-training or finetuning of \llms.
    Second, it can be implemented with a small number of \llms over the decision making process.
    In our experiment, we show that it calls \llms in only $1.5$\% of the time steps.

    \item \textbf{Plug-and-play compatibility.}
    Our framework can flexibly incorporate off-the-shelf pretrained \llms 
    in a plug-and-play manner. 
    {Furthermore, unlike existing methods that require advanced models such as GPT-4 \citep{krishnamurthy2024can}, our approach can leverage much smaller language models (e.g., a model with 80 million parameters) and achieve promising decision making results.}
\end{itemize}

\paragraph{Paper organization.}
The rest of this paper is organized as follows.
In \cref{sec:problem_setting}, we formally introduce sequential decision making settings and characteristics of large language models (\llms). 
In \cref{sec:method}, we introduce our framework for efficiently incorporating \llms into sequential decision making, and discuss its statistical and computational efficiency.
We conduct extensive experiments in \cref{sec:experiment} to verify the effectiveness of our proposed methods, and provide analyses in \cref{sec:analyses}. 
We discuss related work in \cref{sec:related_work} and conclude our paper in \cref{sec:conclusion}.

\section{Problem Setting}
\label{sec:problem_setting}

We focus on contextual bandits, a key problem in sequential decision making that emphasizes the fundamental challenge of balancing exploration and exploitation \citep{lattimore2020bandit}.
In contextual bandits, a learner interacts with an unknown environment over $T \in \bbN^+$ rounds.
At each round $t \in [T]$, the learner receives a context $x_t\in\mathcal{X}$ (the context space), selects an action $a_t\in\mathcal{A}$ (the action space), and then observes a bounded loss $\ell_t (a_t)$ (sampled from an unknown distribution), where $\ell_t: \cA \rightarrow [0,1]$ is the underlying loss function.
Contextual bandits can be viewed as the simplest form of reinforcement learning where state transitions are abstracted away.
Following the convention \citep{agarwal2012contextual, foster2018practical, foster2020beyond}, we assume that the learner has access to a function class $\mathcal{F}\subseteq(\mathcal{X}\times\mathcal{A}\to[0,1])$ to approximate an unknown true loss function $f^\star(x,a) = \E[\ell_t \mid x_t=x, a_t=a]$.
Let $\pi^\star(x) = \argmin_a f^\star(x,a)$ denote the optimal policy with respect to the true expected loss (i.e., always selecting an action that achieves the smallest expected loss).
The learner's goal is to choose a policy $\pi = \prn{\pi_1, \cdots, \pi_T}$ to minimize the cumulative regret, which is defined as  
${\reg}(T) \ldef \sum_{t=1}^{T} f^\star(x_t, \pi_t(x_t)) -  f^\star(x_t, \pi^\star(x_t)) $.

We focus on the setting where the context space and the action space are subspaces of the language space, i.e., the learner interacts with an environment through textual contexts and actions, and actions that induce low loses are usually consistent with commonsense knowledge and/or reasoning.\footnote{
One can also prompt \llms with numerical representations to get regression-style predictions \citep{garg2022can}.}
Therefore, our setting motivates leveraging pretrained large language models (\llms) into contextual bandits.
Specifically, we consider a pretrained \llm: prompt $p$ $\mapsto$ output $o$, that maps a prompt $p$ to a textual response $o$ \citep{brown2020language, ouyang2022training, achiam2023gpt}.
Since \llms are pretrained to acquire general knowledge about the world, we expect the output $o_t \sim \llm(p = x_t)$ of \llms, when prompting \llms with the context $x_t$ (and other relevant information), would provide informative guide for the decision making process.

\paragraph{Additional notation.}
For an integer $n\in\bbN$, we let $[n]$ denote the set
        $\{1,\dots,n\}$.  
        For a finite set $\cZ$, 
        we let $\unif(\cZ)$ denote the uniform distribution over all the
        elements in $\cZ$.
         We use $e_i \in \bbR^d$ to denote the $i$-th canonical vector in $\bbR^d$, i.e., its $i$-th entry is $1$ and the rest entries are $0$.

\section{Methods}
\label{sec:method}

We present our approach for efficiently incorporating \llms into contextual bandits/decision making in this section.
We provide the algorithmic foundation in \cref{sec:framework} and various sampling strategies in \cref{sec:framework_probability}.

\subsection{Efficient Decision Making with \llms}
\label{sec:framework}

At a high level, our framework utilizes an online model selection algorithm
to adaptively balance the performance of two sets of base algorithms: (i) standard contextual bandit algorithms, and (ii) policies constructed based on off-the-shelf pretrained \llms.
Our framework achieves the best-of-both-worlds by (i) efficiently extracting knowledge stored in pretrained \llms and (ii) leveraging the long-term learning ability of standard contextual bandit algorithms.
We construct \llm-powered policies in \cref{sec:framework_llm_policy} and introduce the algorithmic framework in \cref{sec:framework_algorithm}.

\subsubsection{\llms as Decision Making Agents}
\label{sec:framework_llm_policy}

Since the outputs of \llms are in the general language space that may not align with any action in the action set, we first provide an algorithm to convert pretrained \llms to decision making agents.

\begin{algorithm}[t]
	\caption{Construct \llm-Powered Policies}
	\label{alg:llm_policy} 
	\renewcommand{\algorithmicrequire}{\textbf{Input:}}
	\renewcommand{\algorithmicensure}{\textbf{Output:}}
	\newcommand{\algorithmicbreak}{\textbf{break}}
    \newcommand{\BREAK}{\STATE \algorithmicbreak}
	\begin{algorithmic}[1]
\REQUIRE 
Context $x$, pretrained \llm, embedding model $g: \text{language} \rightarrow \bbR^d$, similarity measure $\similarity: \bbR^d \times \bbR^d \rightarrow \bbR$, and hyperparameter $k \in \bbN^+$.
  \STATE Prompt \llm with context $x$ to obtain top-$k$ most likely outputs $o_{{i}}$ and likelihood $q_{{i}}$: $\crl{\prn{o_{{1}}, q_{{1}}}, \cdots, \prn{o_{{k}}, q_{{k}}}}$.\label{line:prompt}
  \STATE 
  Embed all actions $\crl{g(a): a \in \cA} \subseteq \bbR^d$ and \llm outputs $\crl{g(o_{i}) : i \in [k]} \subseteq \bbR^d$.
  \STATE Get ${a_i \ldef \argmax_{a \in \cA} \similarity(g(o_{i}), g(a))}$ for each $i \in [k]$.
  \STATE Construct $\pi^{\llmmath}$ such that $\bbP \prn{\pi^{\llmmath}(x) = a_{{i}}} = q_{{i}} / \sum_{j = 1}^k q_{{j}}$. \label{line:weighted_prob}
	\end{algorithmic}
\end{algorithm}
\cref{alg:llm_policy} is designed to be compatible with flexible choices of \llms, embedding models, and similarity measures.
It prompts the \llm with context $x$  to obtain top-$k$ most likely outputs $o_{{i}}$ and together with their likelihood $q_{{i}}$: $\crl{\prn{o_{{1}}, q_{{1}}}, \cdots, \prn{o_{{k}}, q_{{k}}}}$.\footnote{Additional instructions or prior interaction history can also be incorporated into the prompt design. When $k=1$, we only need to obtain the \llm output $o$, without computing the likelihood $q$.}
For each embedded output $g(o_{{i}})$, it then measures its similarity between each of the embedded action $\crl{g(a), a \in \cA}$, and find the one $a_{i}$ with the highest similarity. 
Finally, we construct policy $\pi^{\llmmath}$ by 
mapping $x$ into the (multi) set $\crl{a_{{1}}, \cdots, a_{{k}}}$ with weighted probability, i.e., 
$\bbP \prn{\pi^{\llmmath}(x) = a_{{i}}} = q_{{i}} / \sum_{j = 1}^k q_{{j}}$.
The \llm-powered policy uses the same policy $\pi^{\llmmath}$ for the entire decision making process to avoid expensive re-training/finetuning of \llms.

\subsubsection{Algorithmic Framework}
\label{sec:framework_algorithm}

\begin{algorithm}[t]
	\caption{Efficient Decision Making with \llms}
	\label{alg:algorithm} 
	\renewcommand{\algorithmicrequire}{\textbf{Input:}}
	\renewcommand{\algorithmicensure}{\textbf{Output:}}
	\newcommand{\algorithmicbreak}{\textbf{break}}
    \newcommand{\BREAK}{\STATE \algorithmicbreak}
	\begin{algorithmic}[1]
\REQUIRE 
A set of contextual bandit algorithms $\crl{\alg^{\cb_{1}}, \cdots, \alg^{\cb_{M_1}}}$, 
a set of \llms $\crl{\llmmath_{M_1+1}, \cdots, \llmmath_{M}}$.
\STATE 
Convert \llms to $\crl{\pi^{\llmmath_{M_1+1}}, \cdots, \pi^{\llmmath_{M}}}$ using \cref{alg:llm_policy}.
  \STATE Order all policies as $\crl{\pi^i}_{i=1}^M$. Initialize sampling strategy $p_1 = \unif{[M]}$.
  \FOR{$t = 1, 2, \dots, T$}
  \STATE Receive contaxt $x_t$.
  \STATE Sample $i_t \sim p_t$. 
  \STATE Follow $\alg^{i_t}$ to play action $a_t$ and observe loss $\ell_t(a_t)$.
  \STATE Update contextual bandit algorithms with $\prn{x_t, a_t, \ell_t(a_t)}$. 
  \STATE Update sampling strategy $p_{t+1} \leftarrow p_t$.\\
  \algcommentlight{We discuss detailed sampling strategies updates in \cref{sec:framework_probability}.} \label{line:update}
  \ENDFOR
	\end{algorithmic}
\end{algorithm}
In \cref{alg:algorithm}, we present our framework to efficiently incorporate \llms into contextual bandits.
\cref{alg:algorithm} leverages online model (expert) selection algorithms \citep{auer2002nonstochastic, agarwal2017corralling, pacchiano2020model} to adaptively balance standard contextual bandit algorithms and \llm-powered policies. 
Compared to existing online model selection algorithms, 
\cref{alg:algorithm} additionally (i) incorporates \cref{alg:llm_policy} to convert \llms into policies, and (ii) allows more flexible sampling strategies to control the number of \llm calls (see \cref{sec:framework_probability} for detailed discussion).
At a high-level, the sampling probability in \cref{alg:algorithm} is designed to rely more on the set of \llm-powered policies at the beginning, and then gradually transit to put more probability on standard contextual bandit algorithms.
By doing so, we aim to simultaneously achieve the following two objectives:
\begin{itemize}
    \item \textbf{Leveraging knowledge in \llms.} At the beginning stage, we leverage \llms to select more informative data to warm start the learning process, and help contextual bandit algorithms learn better.
    \item \textbf{Long-term adaptation to environments.} In the later stage, we leverage the long-term learning ability of contextual bandit algorithms to minimize losses in the long run.
\end{itemize}

\subsection{Sampling Strategies}
\label{sec:framework_probability}
In this section, we discuss in detail how to update the sampling strategy in \cref{alg:algorithm} (line \ref{line:update}).
We present simple, pre-determined sampling strategies in \cref{sec:framework_probability_simple} and learning-based sampling strategies in \cref{sec:framework_probability_corral}.

\subsubsection{Simple Pre-Determined Sampling Strategies}
\label{sec:framework_probability_simple}
We provide several simple, pre-determined sampling strategies in this section. They are simple and can be implemented without additional computation overhead.
They follow the basic idea of putting more probability on \llm-powered policies at the beginning and gradually transiting probability to standard contextual bandit algorithms.
We use $p^{\llmmath}_t$ to denote the total probability of sampling \llm-powered policies, and use $p^\cb_t \ldef 1 - p_t^\llmmath$ to denote the total probability of sampling standard contextual bandit algorithms.
In the following, we focus primarily on updating $p^\llmmath_t$ (and thus $p^\cb_t$).\footnote{One can apply simple strategies (e.g., uniform allocation) to allocate $p^\llmmath_t$ (and $p^\cb_t$) to individual policies.}
We set $0 \leq p_{\min} \leq p_{\max} \leq 1$ as user-specified lower and upper bound on $p^\llmmath_t$.
\begin{itemize} 
    \item \textbf{Polynomial decay.}
    Let $C_{\mathsf{poly}}$ and $\alpha$ be two hyperparameters.
    We set 
    $$p^\llmmath_t \ldef \min\crl{p_{\max}, \max \crl{ p_{\min}, C_{\mathsf{poly}} / t^\alpha}}.$$ 
    \item \textbf{Exponential decay.}
    Let $C_{\mathsf{exp}}$ and $\beta$ be two hyperparameters.
    We set 
    $$p^\llmmath_t \ldef \min\crl{p_{\max}, \max \crl{ p_{\min}, C_{\mathsf{exp}}  \exp(-\beta t)}}.$$
\end{itemize}

\paragraph{Number of \llm calls.}
For these simple sampling strategies, it's easy to see the expected number of \llm calls equals to $\sum_{t=1}^T p^\llmmath_t$.
One can also easily tune hyperparameters to
control the number of \llm calls. 

\subsubsection{Learning-Based Sampling Strategies}
\label{sec:framework_probability_corral}

While there exist many other learning-based sampling strategies, we primarily use 
log-barrier online mirror descent (OMD), also known as the \corral update \citep{agarwal2017corralling}, to update the sampling probability with respect to importance-weighted losses incurred by base algorithms.

\begin{algorithm}[t]
	\caption{Log-Barrier-OMD Update \citep{agarwal2017corralling}}
	\label{alg:omd_update} 
	\renewcommand{\algorithmicrequire}{\textbf{Input:}}
	\renewcommand{\algorithmicensure}{\textbf{Output:}}
	\newcommand{\algorithmicbreak}{\textbf{break}}
    \newcommand{\BREAK}{\STATE \algorithmicbreak}
	\begin{algorithmic}[1]
\REQUIRE 
Learning rate $\eta > 0$, 
previous distribution $p_t$, selected base algorithm index $i_t$, and the incurred loss $\ell_t(a_t)$.
\STATE Construct an importance-weighted loss vector $\wb \ell_t \ldef \frac{\ell_t(a_t)}{p_{t, i_t}} e_{i_t} \in \bbR^{M}$. \label{line:importance_weight}
\STATE Find a constant $\lambda \in \brk{\min_{i} \wb \ell_{t, i}, \max_{i} \wb \ell_{t, i}}$ such that 
$\sum_{i=1}^M \frac{1}{\frac{1}{p_{t, i}} + \eta \prn{\wb \ell_{t, i} - \lambda}} = 1$. \label{line:normalization}
\STATE Return an updated distribution $p_{t+1}$ such that $\frac{1}{p_{t+1, i}} = \frac{1}{p_{t, i}} + \eta \prn{\wb \ell_{t, i} - \lambda}$. \label{line:omd}
	\end{algorithmic}
\end{algorithm}

\cref{alg:omd_update} takes as input an initial learning rate $\eta > 0$, previous sampling distribution $p_t$, the index $i_t$ of selected base algorithm, and the incurred loss $\ell_t(a_t)$.
\cref{alg:omd_update} first constructs the standard importance-weighted unbiased loss estimator for all base algorithms (line \ref{line:importance_weight}), 
and then follow log-barrier online mirror descent to update the sampling distribution with respect to the losses (line \ref{line:omd}).
The update requires a normalization constant $\lambda$ (line \ref{line:normalization}), which can be approximated with numerical root-finding algorithms such as the Brent's method \citep{zhang2011improvement}.

We sample from a smoothed version $\wb p_t$ of the sampling distribution $p_t$ to help contextual bandit base algorithms explore at the beginning stage. Specifically, we clip the (total) sampling probability on \llms $p^{\llmmath}_t$ to $1 - p_{\min}$ if the (total) sampling probability on contextual bandits $p^{\cb}_t$ falls below $p_{\min}$, a user-specified hyperparameter.

\paragraph{Number of \llm calls.}
To control the number of \llm calls, we can either early stop sampling from \llm-powered policies in \cref{alg:algorithm} once the budget $B$ is used up, or further modify the sampling strategy as 
\begin{align}
\label{eq:limit_budget}
\check p^\llmmath_t \ldef \wb p^\llmmath_t \cdot \prn*{\frac{B - N_t}{B}}, \,\check p^\cb_t \ldef 1 - \check p^\llmmath_t,
\end{align}
where $N_t$ represents the number of \llm calls used up to time step $t$.
Both approaches limit the number of \llm calls to at most $B$.

\section{Empirical Results}
\label{sec:experiment}
We conduct extensive experiments to examine the effectiveness of our proposed framework. 
We present experimental setups in \cref{sec:experiment_setup}, our main results in \cref{sec:experiment_result}, and ablation study in \cref{sec:experiment_ablation}.
We defer additional experimental details to \cref{appendix:experiment}.
{Code to reproduce all results is available
at \url{https://github.com/dchen48/DMwithLLM}.}

\subsection{Experimental Setups}
\label{sec:experiment_setup}

\paragraph{Datasets.}
We conduct experiments on two textual contextual bandit datasets, whose details are summarized in \cref{table:datasets}.
\oneshotwiki \citep{singh12:wiki-links,oneshotwiki} is a
named-entity recognition task where contexts are text phrases
preceding and following the mention text, and actions are text
phrases corresponding to the concept names.  
\amazon \citep{Bhatia16} is an extreme multi-label dataset whose
contexts are text phrases corresponding to the title and content
of an item, and actions are integers corresponding to item tags.
We construct binary loss for each dataset, where selecting the correct actions leads to a loss of $0$, and incorrect actions results in a loss of $1$.
In our experiments, we process data in batches with a batch size of $32$.

\begin{table}[h]
\caption{Datasets used for experiments.}
\label{table:datasets}
\centering
\begin{tabular}{c c c }
\toprule
Dataset & $T$ & $|\cA|$  \\
\midrule
\oneshotwiki & $622000$ & $311$  \\
\amazon & $1186239$ & $13330$  \\
\bottomrule
\end{tabular}
\end{table}

\paragraph{Baselines.}
We use \spannerGreedy \citep{zhu2022large} as the contextual bandit baseline, which is an efficient algorithm for textual decision making. 
We use \cref{alg:llm_policy} with $k = 1$ to construct the \llm-powered policy baselines.
We consider various \llm backbones, including \flan \citep{chung2024scaling}, with sizes small (80M parameters), base (250M parameters) and large (780M parameters), and more recent models \gemmamodel (instruct) \citep{team2024gemma} and \gptmodel \citep{gpt4omini}. 

We implement our \cref{alg:algorithm} by combining the two types of baselines mentioned above.
In most of our experiments, we select \llm backbones from the \flan model series;\footnote{Our goal is not to examine the most advanced \llms or contextual bandit algorithms. Instead, we aim to verify that \cref{alg:algorithm} can effectively balance contextual bandit algorithms and \llms policies, and outperform both of them when applied individually.} 
we run additional experiments with \gemmamodel and \gptmodel to verify the efficacy of our method when implemented with more advanced models.
Unless otherwise noted, we implement \cref{alg:algorithm} using \cref{alg:omd_update} and a smoothing parameter $p_{\min} = 0.2$.

\paragraph{Evaluation metrics.}
We evaluate algorithms in terms of both statistical and computational performances. 
Statistically, 
following the convention in contextual bandits, 
we measure the performance in terms of the (average) reward, where one can easily convert loss into reward $r_t(a_t) \ldef 1 - \ell_t(a_t)$.
Computationally, 
since models used in contextual bandit algorithms are relatively lightweight (we empirically verify this in \cref{sec:experiment_result}), we measure the performance in terms of the number of \llm calls.
Our results are averaged over 5 random runs;
shaded area in figures represents the standard error of the mean.

\subsection{Main Results}
\label{sec:experiment_result}

\paragraph{Statistical efficiency.}

\begin{figure}[t]
    \centering
    \includegraphics[width=0.5\linewidth]{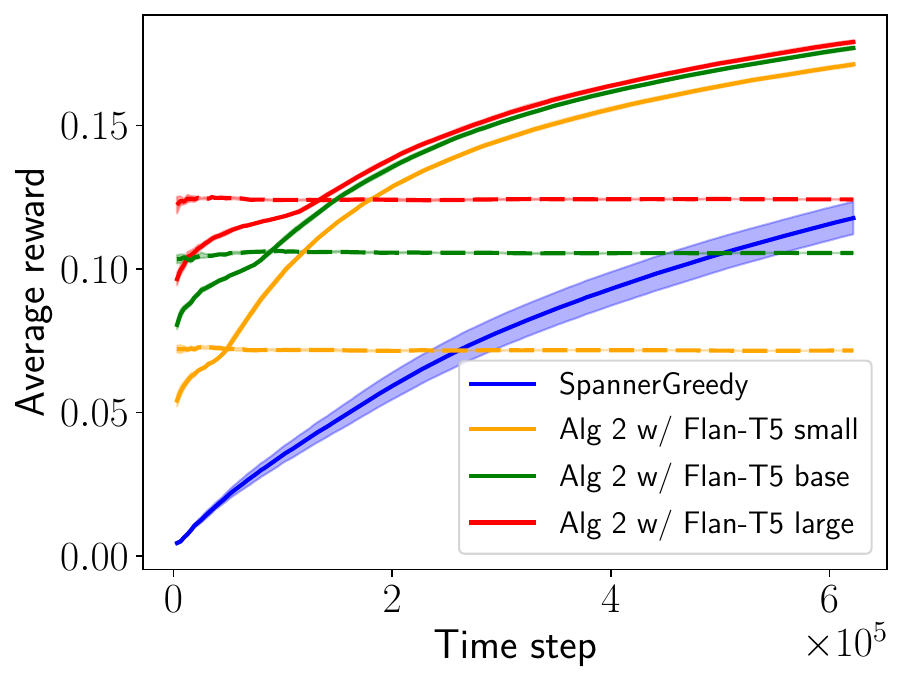}
    \caption{
    Comparison of average reward on the \oneshotwiki dataset (higher is better).
    Our \cref{alg:algorithm} is implemented with various sizes of \flan model. The dashed lines represent the performance of directly applying \llm-powered policy $\pi^{\flan}$ (\cref{alg:llm_policy}) of corresponding sizes.
    }
    \label{fig:oneshotwiki}
\end{figure}

\cref{fig:oneshotwiki} presents comparison of average reward on the \oneshotwiki dataset.
Our \cref{alg:algorithm} significantly outperforms other baselines:
it achieves reward no smaller than $0.17131$ no matter which \flan model is used; on the contrary, even with the \flan large, \llm-powered policy $\pi^\flan$ only achieves reward $0.12423$ and the contextual bandit algorithm \spannerGreedy only achieves reward  $0.11773$. 
The fact the \cref{alg:algorithm} with \flan small (yellow solid line) greatly outperforms $\pi^{\flan\text{-large}}$ (red dashed line) shows the benefits of our algorithmic design. Note \flan small is nearly 10x smaller in parameter count compared to \flan large.

\paragraph{Computational efficiency.}
To examine the computational efficiency, we first run experiments to compare the cost of $\pi^{\llm}$ selection versus the cost of contextual bandit selection, in terms of the execution time. 
As shown in \cref{table:cost}, all $\pi^{\llm}$ selections are considerably more expensive (from $52$x to $159$x more execution time) compared to contextual bandit selection.

\begin{table}[t]
\caption{Cost ratio of $\pi^{\llm}$ selection and contextual bandit selection, measure as the execution time of \flan models divided by the execution time of \spannerGreedy.
}
\label{table:cost}
\centering
\begin{tabular}{c c c}
\toprule
Small (80M)& Base (250M) & Large (780M)\\
\midrule
$52.16$& $79.49$ &$159.20$ \\
\bottomrule
\end{tabular}
\end{table}

\cref{table:ratio_calls} presents the fraction of \llms calls in \cref{alg:algorithm} over the decision making process. \cref{alg:algorithm} not only achieves higher reward (\cref{fig:oneshotwiki}), but also only calls LLMs in a small fraction (from $6$\% to $14$\%) of time steps. 
For comparison, directly applying $\pi^{\flan}$ calls \llm at every time step.

\begin{table}[t]
\caption{Fraction of \llm calls in \cref{alg:algorithm} over the decision making process with \flan models and on the \oneshotwiki dataset.}
\label{table:ratio_calls}
\centering
\begin{tabular}{c c c}
\toprule
Small (80M)& Base (250M) & Large (780M)\\
\midrule
0.06177
&$0.10033$  &$0.14381$ \\
\bottomrule
\end{tabular}
\end{table}

To further improve computational efficiency, we apply \cref{eq:limit_budget} or early stopping to limit the number of \llm calls of our algorithm, and show results in \cref{table:limit_budet}.
Our results show that \cref{alg:algorithm} achieves slightly worse reward when limited to a smaller number of \llm calls.
However, \cref{alg:algorithm} still outperform both baselines with an upper bound $B = 10000$ on the number of \llm calls, which is around $9$x smaller compared to the number of \llm calls used in the unconstrained version of \cref{alg:algorithm}.
\begin{table}[t]
\caption{Limit the number of \llm calls in \cref{alg:algorithm}. 
Experiments conducted with the \flan large model and on the \oneshotwiki dataset. 
}
\label{table:limit_budet}
\centering
\begin{tabular}{ccc} 
\toprule
Algorithms            & \# \llm calls                                   & Reward                                       \\ 
\midrule
\spannerGreedy                    & N/A                                       & $0.11773$                                      \\
$\pi^{\flan \text{-large}}$                    &    $622000$                                  &  $0.12423$                                               \\
\cref{alg:algorithm}         &     $89448.6$                                           &          $0.17913$                                    \\ 
\midrule
\cref{alg:algorithm}        & \multicolumn{1}{l}{\multirow{2}{*}{\# \llm calls}} & \multicolumn{1}{l}{\multirow{2}{*}{Reward}}  \\
w/ \cref{eq:limit_budget} & \multicolumn{1}{l}{}                           & \multicolumn{1}{l}{}                         \\ 
\midrule
$B = 10K$                     & $7669.2$                                         & $0.16836$                                      \\
$B = 20K$                     & $12084.4$                                        & $0.17309$                                      \\
\midrule
\cref{alg:algorithm} w/        & \multicolumn{1}{l}{\multirow{2}{*}{\# \llm calls}} & \multicolumn{1}{l}{\multirow{2}{*}{Reward}}  \\
early stopping & \multicolumn{1}{l}{}                           & \multicolumn{1}{l}{}                         \\ 

\midrule
$B = 10K$                     & $10000$                                         & $0.16774$                                 \\
$B = 20K$                     & $20000$                                        & $0.17508$                                     \\

\bottomrule
\end{tabular}
\end{table}

\begin{figure}[t]
    \centering
    \includegraphics[width=\textwidth]{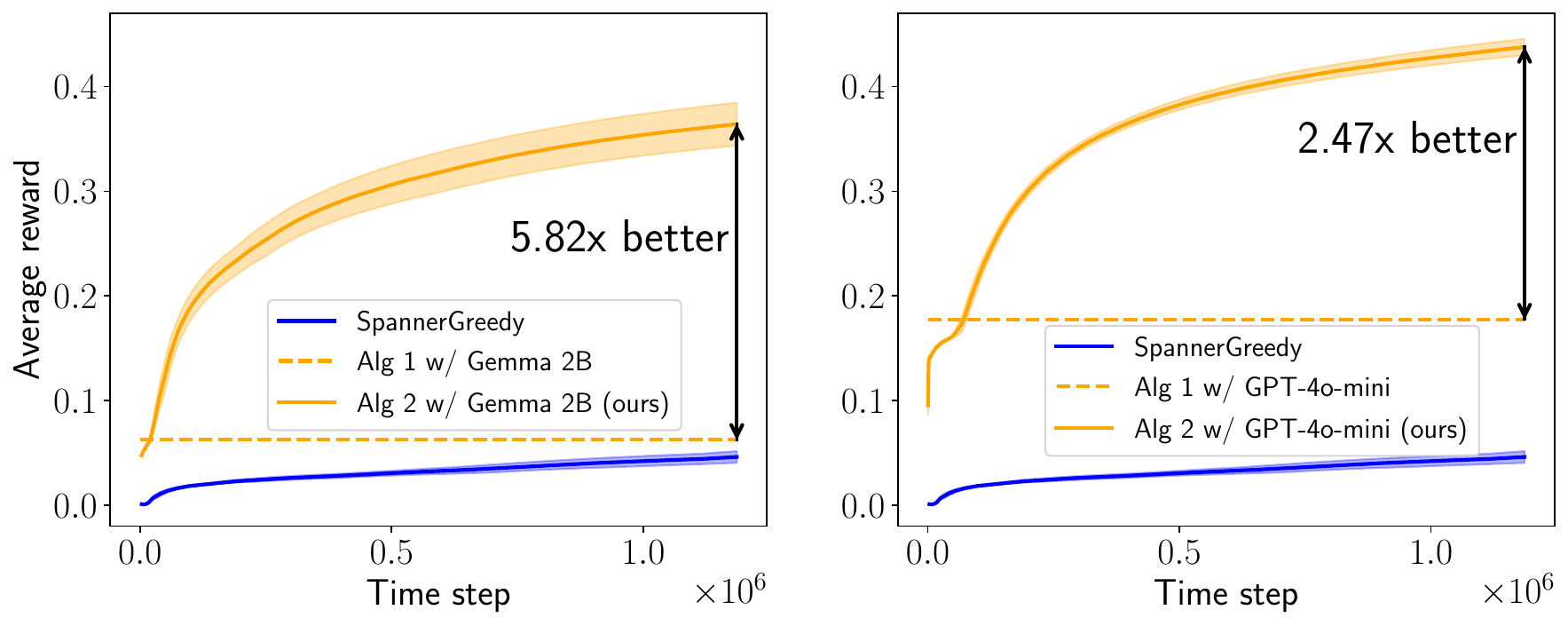}
    \caption{
    Performance comparison (higher is better) on the \amazon dataset. 
    \cref{alg:algorithm} incorporates \spannerGreedy and \cref{alg:llm_policy} as base algorithms.
    \emph{Left:} Use \gemmamodel as the \llm backbone. 
    \emph{Right:} Use \gptmodel as the \llm backbone. 
    }
    \label{fig:advanced_models}
\end{figure}

\paragraph{Large-scale exhibition.}

We conduct a large-scale experiment on the \amazon dataset that has more than $13$k actions (around $42$x larger than the \oneshotwiki dataset). 
With \flan small model, 
as shown in \cref{fig:amazon}, our \cref{alg:algorithm} achieves more than a $6$x performance gain over baselines:
our algorithm achieves reward $0.33603$, yet both \spannerGreedy and $\pi^{\flan}$ achieves reward below $0.05424$.
\cref{alg:algorithm} calls \llms in only $1.5$\% of the time steps ($17783.4$ \llm calls on average over horizon $1186239$).

\paragraph{Learning with more advanced \llms.}

We run additional experiments on the \amazon dataset with more advanced \llms: \gemmamodel and \gptmodel.
We present the results in \cref{fig:advanced_models}.
When using \gemmamodel as the \llm backbone (left plot in \cref{fig:advanced_models}), compared to baselines,  our \cref{alg:algorithm} achieves a 5.82x performance gain and calls \gemmamodel in only {$1.62$\%} of the time steps.
When using \gptmodel as the \llm backbone (right plot in \cref{fig:advanced_models}), compared to baselines,  our \cref{alg:algorithm} achieves a 2.47x performance gain and calls \gptmodel in only {$4.49$\%} of the time steps.\footnote{Due to computational constraints, we calculate the performance of \cref{alg:llm_policy} with \gemmamodel or \gptmodel as the average performance over the first {$96000$} time steps (the first {$3000$} data batches with a batch size {$32$}). These averaged performances should be fairly accurate, as demonstrated by the real-time average performance of the \flan small model in \cref{fig:amazon}, which appears to follow a nearly straight line.}
These results show that our \cref{alg:algorithm} not only works well with \flan models but also with more advanced models such as \gemmamodel and \gptmodel, highlighting the broad compatibility of our algorithmic design.

\subsection{Ablation Study}
\label{sec:experiment_ablation}

\paragraph{Probability updating strategies.}
We examine the performance of various probability updating strategies introduced in \cref{sec:framework_probability}.
Beyond the log-barrier OMD update, we also include simple pre-determined updating strategies: polynomial decay and exponential decay (we set $p_{\min} = 0$ and $p_{\max} = 0.8$). 
For polynomial decay, we set $\alpha = 1$ and select a $C_{\mathsf{poly}}$ from set $\crl{1, 10, 100}$ that achieves the highest reward.
For exponential decay, we select $\beta \in \crl{0.1, 0.01}$ and $C_{\mathsf{exp}} \in \crl{1, 10, 100}$ jointly that achieves the highest reward. 
\cref{table:update_strategies} shows the results of various probability updating strategies: 
while log-barrier OMD achieves better reward, pre-determined updating strategies generally leads to a smaller number of \llm calls.

\begin{table}[h]
\caption{Comparison of different probability updating strategies. Experiments conducted with the \flan large model and on the \oneshotwiki dataset. We record the final average reward.}
\label{table:update_strategies}
\centering
\begin{tabular}{c c c}
\toprule
Methods & \# \llm calls & Reward \\
\midrule
Polynomial decay & $19419.8$ & $0.17413$\\
Exponential decay & $14943.6$ & $0.17259$\\
Log-barrier OMD & $89448.6$ & $0.17913$\\
\bottomrule
\end{tabular}
\end{table}

\paragraph{Smoothing strategy for \cref{alg:omd_update}.}
In \cref{alg:omd_update}, we adopt the smoothing strategy that clip the (total) sampling probability on \llms $p^{\llmmath}_t$ to $1 - p_{\min}$ if the (total) sampling probability on contextual bandit algorithms $p^{\cb}_t$ falls below $p_{\min}$.
By doing this, we help contextual bandit base algorithms within \cref{alg:algorithm} better adapt to the environment, especially at the beginning stage. 
We compare our clipping-type smoothing strategy with the 
mixing-type smoothing strategy proposed in \citet{agarwal2017corralling}: 
given a smooth parameter $\gamma$, set $\wb p_{t} \ldef \prn{1 - \gamma} \cdot p_{t} + \gamma \cdot \unif{[M]}$.
We present the results in \cref{table:smooth}. 
Our result indicates that smoothing \cref{alg:omd_update} is important and our clipping strategy work betters than the mixing strategy.

\begin{table}[h]
\caption{Comparison of different smoothing strategies for \cref{alg:omd_update}. Experiments conducted with the \flan large model and on the \oneshotwiki dataset. 
}
\label{table:smooth}
\centering
\begin{tabular}{c c c}
\toprule
Methods & \# \llm calls & Reward \\
\midrule
 No smoothing &  $618551.4$& $0.12386$\\
\midrule
Clipping (ours) & \# \llm calls & Reward \\
\midrule
$p_{\min} = 0.1$ & $144201.2$ & $0.17532$\\
$p_{\min} = 0.2$ & $89448.6$ & $0.17913$\\
\midrule
Mixing  & \# \llm calls & Reward \\
\midrule
$\gamma = 0.05$ &$151608.0$ & $0.17449$\\
$\gamma = 0.1$ & $149728.6$ & $0.17288$\\
$\gamma = 0.2$ & $189486.0$ & $0.16691$\\
$\gamma = 0.4$ & $248214.4$ & $0.15862$\\
\bottomrule
\end{tabular}
\end{table}

\section{Analyses}
\label{sec:analyses}

\paragraph{\llms empower contextual bandit algorithms.}
As shown in \cref{fig:oneshotwiki}, \cref{alg:algorithm} consistently outperforms its base algorithms.
Since the \llm backbones in \llm-powered policies are never updated (for efficiency reasons), we hypothesizes that our \cref{alg:algorithm} empowers its bandit base algorithms with the help of \llms.

\begin{figure*}[th]
    \centering
    \includegraphics[width=\textwidth]{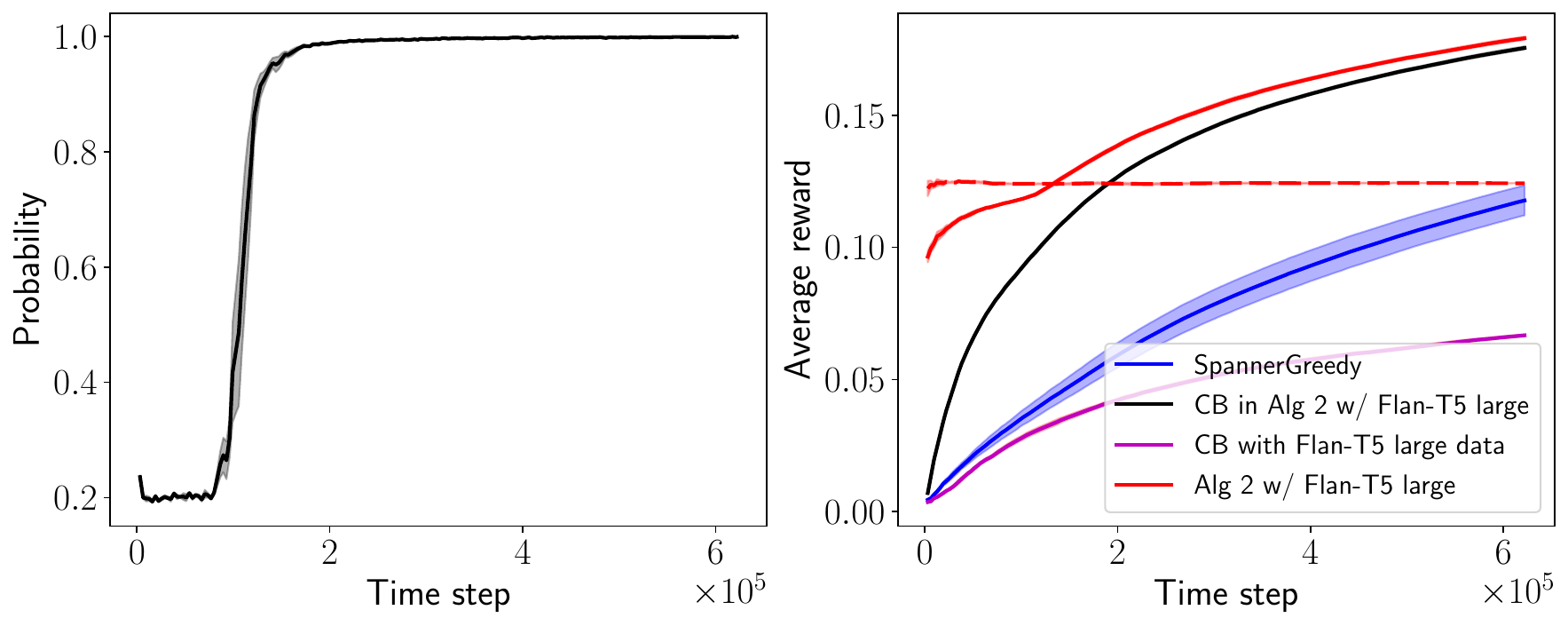}
    \caption{
    Experiments with the \flan large model and on the \oneshotwiki dataset.
    \emph{Left:} Real-time probability $p^{\cb}_t$ of sampling contextual bandit base algorithm in \cref{alg:algorithm}.
    \emph{Right:} Hypothetical performance of the contextual bandit base algorithm within \cref{alg:algorithm} (black solid line) and hypothetical performance of the contextual bandit algorithm learned with purely \llm selected data (solid purple line).
    }
    \label{fig:real_time_prob}
\end{figure*}

To test this hypothesis, we first plot the real-time probability $p^{\cb}_t$ of \cref{alg:algorithm} sampling its contextual bandit base algorithm (left plot in \cref{fig:real_time_prob}). Since $p^{\cb}_t$ quickly increases its value to (around) $1$ after the initial learning stage, we know that the contextual bandit base algorithm within \cref{alg:algorithm} plays an important role after the initial stage. 
We then plot the hypothetical performance of the contextual bandit base algorithm within \cref{alg:algorithm} (as if it were played at every time step). 
As shown in \cref{fig:real_time_prob} (right plot), the contextual bandit base algorithm within \cref{alg:algorithm} (solid black line) achieves much better performance compared to the stand-alone contextual bandit algorithm ($0.17546$ vs. $0.11773$).
Since the main difference lies in the incorporation of data selected by \llm-powered policy,
this shows that \llm selected data helps contextual bandit algorithm learn better.

We also draw the hypothetical performance of \spannerGreedy learned with purely \llm selected data (solid purple line in \cref{fig:real_time_prob} right plot), which is worse than \spannerGreedy ($0.06669$ vs. $0.11773$). This suggests that exploration in contextual bandit algorithm is also important and cannot be replaced with \llm selected data.

\paragraph{\cref{alg:algorithm} with multiple \llms.}
We run \cref{alg:algorithm} with two \llms: \flan large and \flan small.
We compare this approach to \cref{alg:algorithm} with either \flan large or \flan small. 
We use $N_S$ and $N_L$ to denote the number of \flan large and \flan small calls, respectively, and show the results in \cref{table:multiple}.
Compared to learning with a large model, learning with both large and small models achieves slightly worse reward,\footnote{This may be due to the fact that balancing over more models creates larger learning overheads.} but also uses a slightly smaller number of large model calls. 
\cref{alg:algorithm} relies more on the large model (89224 calls) instead of the small model (5833.4 calls on average), as it is designed to automatically adapt to better base policies.

\begin{table}[h]
\caption{\cref{alg:algorithm} with multiple \llms. Experiments conducted on the \oneshotwiki dataset.
}
\label{table:multiple}
\centering
\begin{tabular}{c c c c}
\toprule
\flan models & $N_S$ &  $N_L$ & Reward \\
\midrule
large + small &$5833.4$ &$89224.0$  &$0.17813$\\
large & N/A & $89448.6$ & $0.17913$\\
small &$38424.0$  & N/A &$0.17131$ \\
\bottomrule
\end{tabular}
\end{table}

\section{Related Work}
\label{sec:related_work}

\paragraph{Sequential decision making.}
Sequential decision making is rooted in rich theoretical foundations \citep{langford2007epoch, agarwal2014taming, foster2021statistical}, and
there is a long line of work that develop efficient decision making algorithms with general function approximation \citep{agarwal2012contextual, foster2018practical, foster2020beyond, simchi2021bypassing, zhu2022large, zhu2022contextual, rucker2023infinite, zhang2024efficient}; 
in our experiments, we include one such algorithm to textual environments.
Another line of work focus on developing online model selection algorithms to balance the performance of base algorithms \citep{auer2002nonstochastic, agarwal2017corralling, pacchiano2020model, zhu2020regret, zhu2022pareto,  marinov2021pareto, zhu2022near, dann2024data}. 
Compared to previous online model selection approaches,
we further incorporate \llms into the decision making process.

\paragraph{\llms for decision making.}
While there have been many studies that leverage \llms into supervised learning \citep{xie2021explanation, garg2022can, akyurek2022learning}, the understanding of how to leverage \llms into sequential decision making is less developed.
There exist two main approaches: 
(i) view decision making as sequence modeling and 
pretrain/finetune large models to adapt them to unknown environments \citep{chen2021decision, zheng2022online, reid2022can, sun2023smart, raparthy2023generalization,  lee2024supervised}, and (ii) leverage prompt engineering and in-context learning to adapt pretrained large models to sequential decision making problems \citep{krishnamurthy2024can}.
In this paper, we propose a new approach that 
efficiently incorporates \llms into sequential decision making, addressing drawbacks of previous approaches.

\section{Conclusion}
\label{sec:conclusion}
In this paper, we study the problem of how to efficiently incorporate large language models into contextual bandits, a key problem in sequential decision making that emphasizes the fundamental challenge of balancing exploration and exploitation.
We propose to use online model selection algorithms to adaptively balance \llms agents and standard contextual bandit algorithms.
Statistically, our approach greatly outperforms stand-lone \llm-powered policies and contextual bandit algorithms.
Computationally, our approach avoids the need for expensive re-training or finetuning, and utilizes only a small fraction of \llm calls throughout the decision making process. 
Our framework is highly flexible, allowing for the integration of various off-the-shelf pretrained LLMs. In our experiments, it delivers promising results even when using a language model with only 80 million parameters.

\newpage
\bibliography{refs}

\newpage
\appendix
\onecolumn
\section{Other Details for Experiments}
\label{appendix:experiment}

\subsection{Datasets}

\oneshot \citep{singh12:wiki-links,oneshotwiki} is a
named-entity recognition task where contexts are text phrases (English)
preceding and following the mention text, and actions are text (English)
phrases corresponding to the concept names.  
\oneshotwiki is a subset of this dataset obtained by taking all actions with at least 2000 examples.
We construct binary reward function that is an indicator function for whether the action corresponds to the actual entity mentioned.

\amazon \citep{Bhatia16} is an extreme multi-label dataset whose
contexts are text phrases (English) corresponding to the title and content
of an item, and actions are integers corresponding to item tags.
We construct binary reward function that indicates whether (one of) the correct item tags is selected.

\subsection{Models and Hyperparameters}

\subsubsection{\cref{alg:llm_policy}}

We construct \llm-powered policies using \cref{alg:llm_policy} and various \llm backbones, including \flan models of different sizes \citep{chung2024scaling}, \gemmamodel (instruct) \citep{team2024gemma} and \gptmodel \citep{gpt4omini}.
We use sentence transformer \citep{reimers-2019-sentence-bert} as the embedding model, cosine similarity as the similarity measure, and hyperparameter $k = 1$. 
We provide the prompt design used in line \ref{line:prompt} of \cref{alg:llm_policy} below.

    \paragraph{\oneshotwiki.}
    We only run experiments with \flan models on this dataset.
    Given the text phrases preceding the mention text \verb|text_preceding|, and following the mention text \verb|text_following|, we aim to predict the mention text. 
   Let \verb|<extra_id_0>| represent the masked token in \flan models that needs to be filled in; 
    we construct the prompt as:
\begin{verbatim}
question: text_preceding <extra_id_0>. text_following
\end{verbatim}

\paragraph{\amazon.}
We run experiments with \flan models, \gemmamodel and \gptmodel on this dataset.
Given the \verb|title| and \verb|content| of an item, we aim to predict the associated label.
We construct prompts for different \llms in the following.

\begin{itemize}
    \item \textbf{\flan models.}
    We construct the prompt as:
    \begin{verbatim}
Title: title
Content: content
Task: Predict the associated label.
\end{verbatim}

\item \textbf{\gemmamodel.}
Following the template provided in \citet{gemma2b-huggingface}, we construct the prompt as:
\begin{verbatim}
<bos><start_of_turn>user
Title: title
Content: content
Task: Predict the item tag based on the content and title.<end_of_turn>
<start_of_turn>model
\end{verbatim}

\item \textbf{\gptmodel.} Following the format provided in \citet{openai_tutorial}, we construct system prompt as: 
\begin{verbatim}
   Predict the item tag based on the content and title.
\end{verbatim}
and construct user prompt as:
\begin{verbatim}
   Title: title
   Content: content
\end{verbatim}
\end{itemize}

\subsubsection{Other Models and Hyperparameters}

For \spannerGreedy, we adapt the implementation and hyperparameters from \citet{zhu2022large}.
We use sentence transformer \citep{reimers-2019-sentence-bert} to embed contexts in $\bbR^{1536}$ by concatenating \verb|text_preceding| and \verb|text_following| (\oneshotwiki) or \verb|title| and \verb|content| (\amazon). We use sentence transformer to embed actions in $\bbR^{768}$ and then apply SVD to reduce the dimensionality of actions to $\bbR^{50}$.
\spannerGreedy uses a bilinear function $ f(x, a) = \langle \phi(a), W \phi(x) \rangle $ to make prediction, where $\phi(\cdot)$ represents (pre-processed) embedding for contexts and actions.
For \cref{alg:omd_update}, we set the learning rate $\eta = 0.05$.

\end{document}